# Latent Space Bayesian Optimization with Latent Data Augmentation for Enhanced Exploration

Onur Boyar[1] and Ichiro Takeuchi[1,2*]

[1]Nagoya University, Nagoya, 464-8603, Aichi, Japan.
[2*]RIKEN, Nihonbashi, Chuo-ku, 103-0027, Tokyo, Japan.

*Corresponding author(s). E-mail(s):
ichiro.takeuchi@mae.nagoya-u.ac.jp;
Contributing authors: boyar.onur.g5@s.mail.nagoya-u.ac.jp;

**Abstract**

Latent Space Bayesian Optimization (LSBO) combines generative models, typically Variational Autoencoders (VAE), with Bayesian Optimization (BO) to generate de-novo objects of interest. However, LSBO faces challenges due to the mismatch between the objectives of BO and VAE, resulting in poor exploration capabilities. In this paper, we propose novel contributions to enhance LSBO efficiency and overcome this challenge. We first introduce the concept of latent consistency/inconsistency as a crucial problem in LSBO, arising from the VAE-BO mismatch. To address this, we propose the Latent Consistent Aware-Acquisition Function (LCA-AF) that leverages consistent points in LSBO. Additionally, we present LCA-VAE, a novel VAE method that creates a latent space with increased consistent points through data augmentation in latent space and penalization of latent inconsistencies. Combining LCA-VAE and LCA-AF, we develop LCA-LSBO. Our approach achieves high sample-efficiency and effective exploration, emphasizing the significance of addressing latent consistency through the novel incorporation of data augmentation in latent space within LCA-VAE in LSBO. We showcase the performance of our proposal via de-novo image generation and de-novo chemical design tasks.

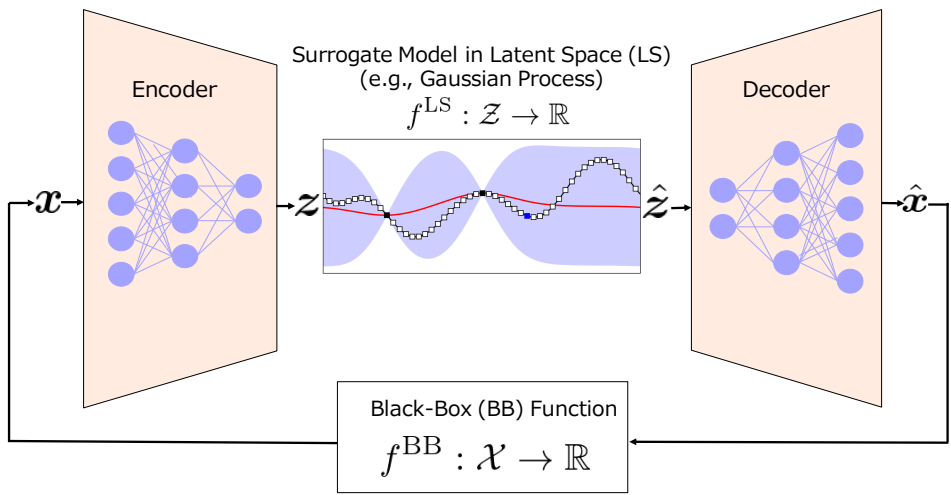

**Fig. 1** An illustration of LSBO. In LSBO, an expensive-to-evaluate BB function optimization problem is solved by BO in latent space constructed by a generative model such as VAE. First, an input instance $x$ in the input domain $\mathcal{X}$ is encoded as a latent variable $z \in \mathcal{Z}$ by the encoder, and the surrogate model (e.g. Gaussian Process) is trained with $z$ and $f^{BB}(x)$. Then, by running a BO in the latent space by using a Bayesian surrogate model in the latent space denoted as $f^{LS}$, a latent variable $\hat{z}$ is suggested by the acquisition function (AF) of the BO, and it is decoded as a new input instance $\hat{x}$. The newly generated input instance $\hat{x}$ is fed into the BB function $f^{BB}$ and obtains its property $\hat{y} \in \mathbb{R}$. This process is repeated in an LSBO. More details are provided in §2.2.

# 1 Introduction

Bayesian Optimization (BO) is used for optimizing expensive-to-evaluate black-box (BB) functions[1]. A key property of BO is to consider the *exploration-exploitation trade-off*, i.e., finding a trade-off between exploring new regions of the search space and exploiting the knowledge gained from previous evaluations. BO is effective when the search space is represented in a low-dimensional vector space. However, applying BO to high-dimensional data or structural data such as chemical compounds is challenging. To overcome this challenge, a method called *Latent Space Bayesian Optimization (LSBO)* has been proposed [1], in which the latent space of the high-dimensional/structure data is identified by using generative models such as Variational Autoencoder (VAE) and BO is performed in the latent space. Figure 1 illustrates LSBO.

Unfortunately, a simple combination of VAE and BO usually does not yield satisfactory results. This can be attributed to the *mismatch* between the objectives of VAE and LSBO. The main objective of VAE is to generate instances that closely resemble the training instances by sampling from the dense region in the latent space. On the other hand, the main focus of the exploration-phase in LSBO is to generate *de-novo* instances that are typically dissimilar to the training instances and are sampled from sparse regions in the latent space. This mismatch results in an off-target low quality generation. Therefore, to enable LSBO to effectively perform and balance exploration and exploitation, it is indispensable to identify a latent space that can resolve this mismatch.

---

[1]List of all abbreviations is provided in Table 6 in the appendix.

In this study, we examine two critical issues in LSBO that stem from the mismatch. The first issue is what we refer to as *latent consistency/inconsistency*. Consider a situation where a single point in the latent space is first decoded and then it is again encoded back into the latent space. In this situation, we refer to a point in the latent space as *latent consistent* when the point obtained by passing through the decoder and encoder of the VAE is located in the same point, whereas we refer to a point as *latent inconsistent* when the two points are located differently[2].

In LSBO, since a BB function evaluation is performed after decoding the evaluation point in the latent space, if the evaluation point is latent inconsistent, the obtained BB function value will no longer match that of the actual evaluation point[3]. To address the issue of latent inconsistency, we propose a new type of acquisition function (AF) for LSBO called *latent consistent-aware AF (LCA-AF)*. The LCA-AF enables us to focus on searching only for latent consistent points in the latent space in LSBO, thereby avoiding the aforementioned latent inconsistency issue.

The second issue is the limited latent consistent points in the latent space, which hinders the exploration of a wide search space in LSBO. When conducting exploration in LSBO, it involves exploring regions where training instances do not exist, namely, the low-density (sparse) regions in the latent space. However, these low-density regions often exhibit latent inconsistency, making it difficult to perform sufficient exploration. In addition, Figure 2(A) illustrates the results of a simple toy experiment in which numerous latent variables migrate towards the center in the latent space where the density is high, and Figure 2(B) demonstrates the density of the latent consistent points in the latent space, which is confined to a narrow region.

We propose a new VAE method in order to address the issue of limited latent consistent points. We refer to this method as the *latent consistent-aware VAE (LCA-VAE)*. The fundamental idea of LCA-VAE is to train a VAE such that the regions to be explored in the latent space have latent consistency by utilizing *Data Augmentations in Latent Space*. Traditional data augmentation strategies operate in *input space*, but when dealing with complex high-dimensional data, augmenting in the low dimensional latent space offers a label-free and efficient alternative as well as a unique combination with the LSBO methodologies. By augmenting instances from regions identified as promising by the LSBO algorithm, we achieve dual benefits. Figures 2(C) and 2(D) indicate the results of the proposed LCA-VAE, indicating that the latent consistency is widened compared with the case of vanilla-VAE in Figs 2(A) and 2(B).

---

[2]This notion is closely related with cycle-consistency, which has been introduced in a different context in [2].

[3]Another strategy to aim this mismatch can be using the latter location in the latent space, and using it in GP rather than the former location in latent space, which is known as *recentering* [3]. This strategy does not solve the problem since the information on the location of the promising region in the latent space is already lost.

The key contributions of our paper are as follows:

1. We introduce the concept of latent consistency/inconsistency as an important problem in existing LSBO. To address the problem, we propose a new AF for LSBO, called *LCA-AF*, which enables exploration in the latent consistent points.
2. We introduce *LCA-VAE*, a novel VAE method that can generate a latent space which contains a larger number of latent consistency points by leveraging *Data Augmentations in Latent Space*, thereby improving BO's exploration capabilities as well as the sample efficiency of LSBO.
3. We combine LCA-VAE and LCA-AF to develop a new LSBO method, which we call, *Latent Consistency Aware-LSBO (LCA-LSBO)*. Through image generation and de-novo chemical design experiments, we validate the improved performance of LCA-LSBO, demonstrating its enhanced LSBO capabilities.

## 1.1 Related Works

The VAE [4] is one of the most popular generative models with many applications such as image generation [5], anomaly detection [6], and denoising [7]. The BO has also been intensively studied as a BB function optimization method, particularly for problems where the evaluation cost of BB functions is high [8]. The LSBO was first introduced for *de-novo molecular design* in [1]. Then, the LSBO was applied to various problems such as neural architecture search [9], image generation [10], material discovery [11], and protein design [12].

Many studies have been conducted to improve the LSBO method. First, in the seminal work in [1], the authors first proposed a method that simply combines the standard VAE and the standard BO, which we refer to as the *vanilla-VAE*. Additionally, they proposed an approach to make the latent space more suitable for label prediction by performing label (e.g., chemical property) prediction on the latent space during VAE training, which we refer to as the *predictor-VAE*. As an other approach to incorporate label information into LSBO, Richards and Groener [13] suggested to use conditional $\beta$-VAE. To improve the validity of instances generated by LSBO, Griffiths and Hernández-Lobato [14] proposed a specific method in the context of molecular design problems. Siivola et al. [15] conducted a study to evaluate how factors such as the dimension of the latent space and the choice of acquisition function affect the performances of LSBO.

The above methods are still sub-optimal because the new information obtained through BO iteration(s) is not incorporated into the VAE. Tripp et al. [16] proposed a weighted retraining approach where the VAE is retrained by an updated training set after BO iteration(s), where the weights of the instances in the updated training set are set to be proportional to their label information. Taking a different direction, Maus et al. [3] tackles the problem of high-dimensional search space of LSBO problems and proposes a local optimization based LSBO approach, LOL-BO. Furthermore, Grosnit et al. [17] introduced a metric learning approach while retraining the VAE. Berthelot et al. [10] proposed invariant data augmentations to improve the quality of the latent space and make it more amenable to BO.

Finally, let us mention several existing studies that have been discussed in different contexts than the LSBO but are relevant to our proposed LCA-LSBO method. The similar notion of latent consistency has been discussed for enhancing the disentanglement of the latent space for image generation task [2, 18]. In the context of BO, several methods have been developed to handle high-dimensional or structured data. In particular, an approach called manifold GP can be interpreted as a method for conducting BO in a latent space [19, 20]. As we describe later, we retrain the VAE by incorporating artificially generated latent variables, which allows us to adapt the latent space for the LSBO task. This approach can also be regarded as a variant of adaptive data augmentation (see, e.g., [21]).

## 2 Preliminaries and Problem Definition

Here, we present preliminary information and problem definition.

### 2.1 VAEs

A VAE consists of an encoder network $f_\phi^{\text{enc}} : \mathcal{X} \to \mathcal{Z}$ and a decoder network $f_\theta^{\text{dec}} : \mathcal{Z} \to \mathcal{X}$ where $\mathcal{X}$ is the domain of input variables $\bm{x}$ and $\mathcal{Z}$ is the domain of latent variables $\bm{z}$, where $\phi$ and $\theta$ are the parameters of the encoder and the decoder, respectively. In VAEs, input variables $\bm{x}$ and latent variables $\bm{z}$ are considered as random variables. Let $p_\theta(\bm{x})$ and $p_\theta(\bm{z})$ be the probability functions of $\bm{x}$ and $\bm{z}$, respectively, where $\theta$ is the set of parameters for characterizing the probabilities[4]. The encoder network $f_\phi^{\text{enc}}$ is formulated as a conditional probability $q_\phi(\bm{z} \mid \bm{x})$, which is considered as an approximation of $p_\theta(\bm{z} \mid \bm{x})$, whereas the decoder network $f_\theta^{\text{dec}}$ is formulated as a conditional probability $p_\theta(\bm{x} \mid \bm{z})$.

A VAE is trained by maximizing

$$J_{\text{VAE}}(\phi, \theta) = \mathbb{E}_{\bm{z} \sim q_\phi(\bm{z}|\bm{x})} \bigl[\log p_\theta(\bm{x} \mid \bm{z}) - \beta D_{\text{KL}}\left(q_\phi(\bm{z} \mid \bm{x}) \| p_\theta(\bm{z})\right)\bigr], \quad (1)$$

where $\mathbb{E}_{\bm{z} \sim q_\phi(\bm{z}|\bm{x})}$ is an expectation operator for random variables $\bm{z}$ that follows the conditional distribution $q_\phi(\bm{z} \mid \bm{x})$, $D_{\text{KL}}(\cdot \| \cdot)$ is Kullback Leibler (KL) divergence between two distributions, and $\beta > 0$ is a hyperparameter to trade-off the balance between the two terms. In (1), the prior distribution of the latent variables $p_\theta(\bm{z})$ is usually set as a multivariate normal distribution $\mathcal{N}(\bm{0}, I)$. When $\beta = 1$, the objective function is reduced to that of the standard VAEs and it can be interpreted as a lower bound of the log likelihood of the input distribution $p_\theta(\bm{x})$. On the other hand, VAEs with $\beta \neq 1$ is called $\beta$-VAE [22].

The conditional probability $q_\phi(\bm{z} \mid \bm{x})$ in the encoder network is formulated as follows. Given an input variable $\bm{x}$, the latent variable $\bm{z}$ is encoded as $\bm{z} = \bm{\mu}_\phi(\bm{x}) + \bm{\sigma}_\phi(\bm{x})\bm{\varepsilon}$, $\bm{\varepsilon} \sim \mathcal{N}(\bm{0}, I)$, where $\bm{\mu}_\phi : \mathcal{X} \to \mathcal{Z}$ and $\sigma_\phi : \mathcal{X} \to \mathbb{R}^+$ represent the mean and the standard deviation of the latent variable corresponding to $\bm{x}$, respectively, and they are implemented together in the encoder network $f_\phi^{\text{enc}}$ with the trainable parameters $\phi$, whereas $\varepsilon$ is a random variable sampled from $\mathcal{N}(\mathbf{0}, I)$. Given a latent variable $z$, using the conditional probability $p_\theta(\boldsymbol{x} \mid \boldsymbol{z})$ in the decoder network $f_\theta^{\text{dec}}$, an input variable $\hat{\boldsymbol{x}}$ is generated as $\hat{\boldsymbol{x}} \sim p_\theta(\boldsymbol{x} \mid \boldsymbol{z})$.

---

[4] Note that the notation $\theta$ is also used as the parameters for the decoder network, and the reason for this will be clarified below.

## 2.2 LSBO

In LSBO, we start with a large number of *unlabeled* instances $\{\boldsymbol{x}_i\}_{i \in [\mathcal{U}]}$ and a small number of *labeled* instances $\{(\boldsymbol{x}_i, y_i)\}_{i \in [\mathcal{L}]}$, where $\boldsymbol{x}_i \in \mathcal{X}$ is an input instance such as a chemical compound and $y_i \in \mathcal{Y} \subseteq \mathbb{R}$ is the label of the input instance $\boldsymbol{x}_i$ such as drug-likeness of the chemical compound $\boldsymbol{x}_i$. Here, the sets of indices of the unlabeled and labeled instances are denoted as $\mathcal{U}$ and $\mathcal{L}$, respectively, and $\mathcal{Y} \subseteq \mathbb{R}$ indicates the space of the label which is assumed to be scalar in this study[5].

A BO is used for a BB function optimization problem when an evaluation of the BB function is expensive (in terms of time, money, etc.). Let us denote the BB function as $f^{\text{BB}} : \mathcal{X} \to \mathcal{Y}$. The goal of a BO is to find $\boldsymbol{x} \in \mathcal{X}$ that maximizes the BB function with as small number of BB function evaluations as possible. The basic idea of BOs is to use a *surrogate model* for the BB function, and it is common to employ a GP model as the surrogate model. If a good surrogate GP model on the input instance space $\mathcal{X}$ can be constructed using the labeled instances $\{(\boldsymbol{x}_i, y_i)\}_{i \in \mathcal{L}}$, the predictive distribution of the label can be obtained for each input instance with unknown labels. Using the predictive distributions, one or more candidates of input objects whose labels are predicted to be greater than the current maximum value $\max_{i \in \mathcal{L}} y_i$ can be selected. After evaluating the BB function for the candidate input objects, the labeled set $\mathcal{L}$ and the GP surrogate model are updated. This process is repeated until we obtain the maximum value or use out the available resource budget.

Unfortunately, it is often difficult to develop a good GP surrogate model on the input instance space $\mathcal{X}$ when $\boldsymbol{x}$ is a high-dimensional complex instance such as a chemical compound. The basic idea of LSBO is to first train a VAE using an unlabeled set $\mathcal{U}$, and then fit a GP surrogate model on the latent space $\mathcal{Z}$ in the trained VAE. Fitting a GP model on the latent space $\mathcal{Z}$ can be easier than fitting it on the input instance space $\mathcal{X}$ because the former is lower-dimensional vector space. Let us denote a GP model in the latent space by $f^{\text{GP}} : \mathcal{Z} \to \mathcal{Y}$ which is fitted by using $\{(f_\phi^{\text{enc}}(\boldsymbol{x}_i), y_i)\}_{i \in \mathcal{L}}$. At each iteration of LSBO, by applying the AF to the predictive distributions obtained by the GP surrogate model on the latent space, the latent variable that maximizes the AF, i.e., $\boldsymbol{z}_{i'} = \operatorname{argmax}_{\boldsymbol{z} \in \mathcal{Z}} f^{\text{AF}}(\boldsymbol{z})$ can be obtained, where $f^{\text{AF}} : \mathcal{Z} \to \mathbb{R}^+$ is an AF of the LSBO. Using the decoder, an input instance $\boldsymbol{x}_{i'} = f_\theta^{\text{dec}}(\boldsymbol{z}_{i'})$ is obtained as the target of the next BB function evaluation. The labeled set is then updated as $\mathcal{L} \leftarrow \mathcal{L} \cup \{i'\}$ and the GP model is also updated accordingly. Here, an optional but useful step is to retrain the VAE model by using the updated dataset $\mathcal{L} \cup \mathcal{U}$ in order to incorporate the new information to the model [16, 17]. As in the standard BO, this process is repeated until we obtain the maximum value or use out the available resource budget. For example, in a chemical compound design problem, the LSBO is used to find a chemical compound with desired property.

---

[5]In this study, although we consider regression problems and $y$ is a scalar variable, we refer to it as a *label* in accordance with active learning terminology.

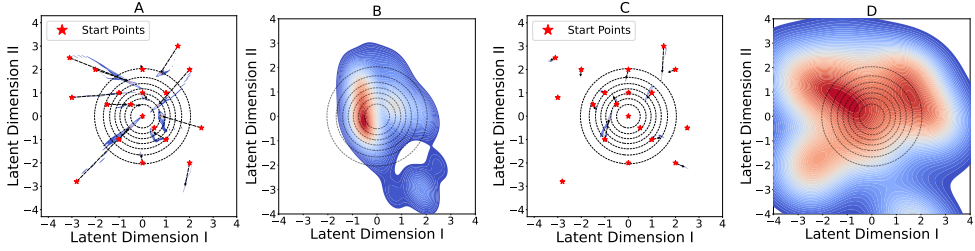

**Fig. 2** Latent consistent points in 2-dimensional latent space of VAE trained on MNIST dataset [23]. **A** and **B** show the latent space of the standard VAE while **C** and **D** show the latent space of proposed LCA-VAE. Dotted circles indicate the contour plot of the prior distribution of the models, $\mathcal{N}(0, I)$. Black dotted arrows in **A** and **C** indicate how the latent variables traverse in the latent space when we repeat encoding and decoding many times. In **A**, many latent variables traverse closer to regions of higher density under $\mathcal{N}(0, I)$, whereas in **C**, it can be observed that most latent variables remain in almost the same position. LSBO behaves as intended only in latent consistent points. Therefore, when the latent consistent points are limited as in **B**, the exploration capability of LSBO is limited. On the other hand, in **D**, the proposed LCA-VAE has expanded the latent consistent points, enabling more extensive exploration.

---

**Algorithm 1** Succesive cycles of $z$

---

**Require:** VAE encoder $f_\phi^{\text{enc}}$, VAE decoder $f_\theta^{\text{dec}}$, Burn-in threshold B, the number of cycles $M$, latent variable $z$
1: Initialize an empty set $Z$
2: **for** $j = 1$ to $M$ **do**
3:      $z^j \leftarrow f_\phi^{\text{enc}}(f_\theta^{\text{dec}}(z))$
4:      **if** $j \geq B$ **then**
5:          Add $z^j$ to $Z$
6:      **end if**
7:      $z \leftarrow z^j$
8: **end for**
9: **return** $Z$

---

### 2.3 Two Critical Issues in LSBO

In this section, we discuss the two critical issues in the current LSBO.

#### 2.3.1 Latent Inconsistency

Given a latent variable $z$, we denote $z^1 = f_\phi^{\text{enc}}(f_\theta^{\text{dec}}(z))$, $z^2 = f_\phi^{\text{enc}}(f_\theta^{\text{dec}}(z^1))$, ..., $z^M = f_\phi^{\text{enc}}(f_\theta^{\text{dec}}(z^{M-1}))$, where each step $f_\phi^{\text{enc}}(f_\theta^{\text{dec}}(z^j)), j \in [M]$ is referred as a *cycle* and the $M$ denotes the number of cycles. For a latent variable $z \in \mathcal{Z}$, we formally say that it is latent consistent if $z = z^M$ when $M \to \infty$ and latent inconsistent otherwise. Note that, for a latent variable $z$ to be consistent, it has to be converged to a single point in the latent space. For latent inconsistent points, we consider the limiting distribution after a sufficient number of cycles. For a sufficiently large $B$, the limiting distribution of a latent variable $z$ is defined as $\{z^j\}_{j=B}^M$, $M \to \infty$. Algorithm

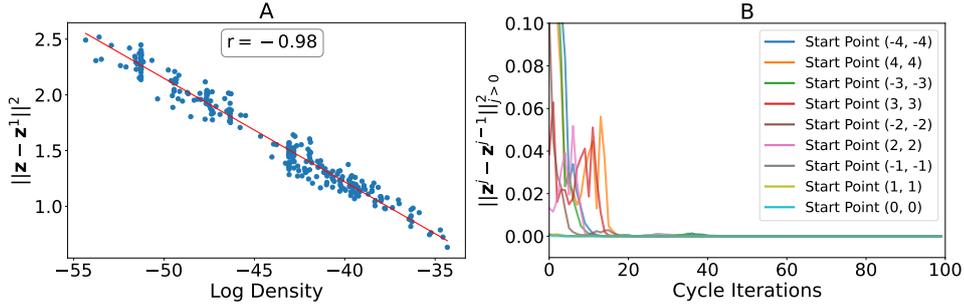

**Fig. 3** Figures demonstrate the relationship between latent inconsistency and the density of the latent variable. **A** shows the $||z - z^1||^2$ values of latent variables sampled from varying densities in latent space. **B** shows the cycles of latent variables from different regions in latent space, where the Y-axis shows the difference between successive cycles, $||z^j - z^{j-1}||^2$, and X-axis shows the cycle iterations. In **A**, inconsistencies increase as the density in latent space decreases. In **B**, convergence requires more cycle iterations as the density in latent space decreases.

1 describes how to find the limiting point/distribution by using the decoder and the encoder of a trained VAE.

When analyzing the latent space, it is observed that the proportion of consistent and inconsistent points, as well as the convergence speed, vary depending on the density of training instances. In a VAE, it is common to set the prior distribution as a standard normal distribution $N(\mathbf{0}, I)$, resulting in high-density near the origin $\mathbf{0}$, which decreases as we move away from the origin. Figure 3(A) illustrates the change in latent inconsistency (measured as $||z - z^1||^2$) with density in $N(\mathbf{0}, I)$. It is evident that the issue of latent inconsistency is more pronounced in low-density regions. Additionally, Fig. 3(B) demonstrates the convergence properties of latent variables in different locations, revealing that latent variables in low-density regions converge slowly.

The important point to note is that the LSBO does not work as intended in regions with latent inconsistencies. In the LSBO process, a new latent variable $z_{\text{new}}$ is selected based on the AF of a BO algorithm. It is then decoded to $x_{\text{new}}$ and its corresponding BB function value $f^{\text{BB}}(x_{\text{new}})$ is obtained. The new information is incorporated into the input dataset, which is then used to retrain both the VAE and the surrogate model. By retraining the models, the new knowledge is effectively integrated into the VAE and the surrogate model. However, a contradiction arises when $z_{\text{new}}$ is latent inconsistent. Specifically, during the update of the surrogate model, the BO algorithm queries the BB function value at $z_{\text{new}}$, but actually obtains the value corresponding to $z^1_{\text{new}} = f^{\text{enc}}\phi(f^{\text{dec}}\theta(z_{\text{new}}))$. This means that the surrogate model in the latent space cannot be updated as intended[6].

Namely, LSBO does not work well in latent inconsistent points, meaning that we must focus on latent consistent points in LSBO. In the next section, we propose the LCA-AF as a new type of AF for LSBO. By utilizing LCA-AF, it becomes possible to focus the search on the regions of latent consistency in the latent space.

---

[6] More discussion on the impact of latent inconsistencies on AF values is provided in the Appendix.

### 2.3.2 Limited Latent Consistent Points

Although LSBO works well when using the limited latent consistent points as shown in Fig. 2(B), it fails to cover a wide range of latent space effectively. Therefore, a search strategy that solely focuses on latent consistent points cannot explore low-density regions. One key aspect of BO is the consideration of the exploration-exploitation trade-off. Exploitation mainly happens in high-density regions where known instances exist, while exploration typically occurs in low-density regions where no existing instances are found. This means that when a search strategy solely concentrates on the latent consistent point, it implies only exploitation is carried out, and exploration of unseen low-density regions is neglected.

Actually, as shown in Figs. 2(A) and (B), latent variables in sparsely populated regions often move towards densely populated regions near the origin. This means that even when we prioritize exploration, we are actually engaging in exploitation. In other words, the LSBO may end up exploring regions that have already been exploited, completely missing out on the opportunity to explore new and unexplored regions of interest. Since the new instance(s) are used to update the VAE to increase the chance of generating the desired object, such latent inconsistencies result in an attempt to update the dense regions. This requires a massive amount of LSBO iterations in order to bring a meaningful update since dense regions include latent representations of training instances, and replacing them with new representations requires many new instances. However, in an LSBO setting, it is unrealistic to have such a high value of new instances through LSBO iterations due to the high cost of experiment in each LSBO step. Therefore, unfortunately, as long as such a restricted search strategy is used, it becomes challenging to discover entirely new instances such as de-novo chemical compounds in an efficient way.

In the following section, we introduce the LCA-VAE as a new method for retraining VAEs. By utilizing this retraining method, we can expand the latent consistent points in the latent space, which in turn widens the search space for the LSBO and enables exploration of previously unseen promising regions. When comparing the latent consistent points in Figs. 2(A) and (B) with those obtained by LCA-VAE in Figs. 2(C) and (D), it is evident that the latter is wider, demonstrating the ability to explore even in low-density regions where there are no training instances available.

## 3 Proposed Method

In this section, we introduce our proposals: LCA-AF, LCA-VAE, and LCA-LSBO. These proposed methods help us to tackle the problem of latent inconsistencies, and the limited consistent points in the latent space, improving the performance of LSBO algorithms by leveraging data augmentation in latent space.

### 3.1 LSBO with Consistent Points: LCA-AF

The idea behind the LCA-AF is to alleviate the disadvantage of latent inconsistencies by taking into account only the available consistent points. In LCA-AF, we modify AFs to use only consistent points in the latent space for LSBO algorithms. After a sufficiently large number of cycles $B$, if we observe $z^{B+1} - z^B \approx z^{B+2} - z^{B+1} \approx, ..., \approx z^{M-1} - z^{M-2} \approx z^M - z^{M-1} \approx \epsilon$, where $\epsilon$ is a constant taking values close to zero, we identify approximate convergence to a consistent point at $z^B$. Since we are interested in the consistent points in LCA-AF, we can discard the latent variables obtained at the earlier cycle iterations than B, before convergence is obtained. Therefore, given a latent variable $z$, our interest lies in $\{f^{\text{AF}}(z^j)\}_{j \in [B,M]}$. The AF values are then computed by taking the mean of the AF values of $\{z^j\}_{j \in [B,M]}$. We denote the resulting AF value as

$$\hat{f}^{\text{LCA-AF}}(z^j)_{j \in [B,M]} = \frac{\sum_{j=B}^{M} f^{\text{AF}}(z^j)}{M - B}. \tag{2}$$

Note that when differences between successive cycles become negligible, we obtain $f^{\text{AF}}(z^B) = f^{\text{AF}}(z^{B+1}) = \cdots = f^{\text{AF}}(z^M)$. This indicates that the limiting value $z^B$ has been identified, and it can be employed as the consistent point. Therefore the Eq. 2 simplifies to the AF value at the cycle iteration B, $\hat{f}^{\text{LCA-AF}}(z^j)_{j \in [B,M]} = f^{\text{AF}}(z^B)$, which we denote as $\hat{f}^{\text{LCA-AF}}(z^B)$[7][8].

We describe an LSBO algorithm that incorporates the LCA-AF in Algorithm 2. This algorithm considers the case where the VAE is retrained after new instance(s) are obtained by BO which uses the proposed AF demonstrated above. Selection by LCA-AF maintains consistency in LSBO, and therefore helps us to deal with the contradictory behavior of LSBO emerges from the latent inconsistencies, enabling LSBO to work as it is supposed to work.

---

**Algorithm 2** LSBO with LCA-AF

**Require:** Unlabeled instances: $\mathcal{U} = \{x_i\}_{i \in [\mathcal{U}]}$, Labeled instances: $\mathcal{L} = \{(x_i, y_i)\}_{i \in [\mathcal{L}]}$, $f^{\text{AF}}(z)$, AF $f^{\text{AF}}$, BB Function $f^{\text{BB}}$, Experiment count $J$, Burn-in threshold $B$, VAE Encoder $f_\phi^{\text{enc}}$ and VAE Decoder $f_\theta^{\text{dec}}$

1: **for** $j = 1$ to $J$ **do**
2:    Fit GP model using $\{(f_\phi^{\text{enc}}(x_i), y_i)\}_{i \in \mathcal{L}}$
3:    Find $z \leftarrow \arg\max \hat{f}^{\text{LCA-AF}}(z^B)$
4:    Generate object: $\hat{x}^* = f_\theta^{\text{dec}}(z)$
5:    Evaluate label: $y^* = f^{\text{BB}}(\hat{x}^*)$
6:    Update: $\mathcal{L} \leftarrow \mathcal{L} \cup (\hat{x}^*, y^*)$
7:    ReTrain $f_\phi^{\text{enc}}$ and $f_\theta^{\text{dec}}$ using $\{\{x_i\}_{i \in [\mathcal{U}]} \cup \{x_i\}_{i \in [\mathcal{L}]}\}$
8: **end for**

---

[7] In high-dimensional latent spaces, even though differences between successive cycles stabilize, we often observe higher differences. In the appendix, we show that these differences are still negligible, therefore allowing us to simplify the LCA-AF calculation in general.

[8] We discuss the optimization details of LCA-AF in the appendix.

## 3.2 LCA-VAE

As discussed in §2.3, the availability of consistent points is limited, particularly in low-density regions of the latent space, limiting the exploration capabilities of BO. To tackle this problem and to enhance the exploration capabilities of LSBO, we introduce a novel VAE model called LCA-VAE. It is designed to improve latent consistencies specifically in these low-density regions and increase the number of consistent points in latent space. This is achieved through the inclusion of the *Latent Consistency Loss (LCL)* in the VAE objective function. For a latent variable $\hat{z} \in \mathcal{Z}$, the LCL is defined as

$$\text{LCL}(\hat{z}) = ||\hat{z} - \hat{z}^1||^2, \tag{3}$$

which calculates the distance between the latent variable $\hat{z}$ and its first cycle $\hat{z}^1 = f_\phi^{\text{enc}}(f_\theta^{\text{dec}}(\hat{z}))$. Using the LCL, the objective function of the LCA-VAE is written as

$$J_{\text{VAE}}^{\text{LCA}}(\phi, \theta) = J_{\text{VAE}}(\phi, \theta) - \gamma \mathbb{E}_{\hat{z} \sim p_{\text{ref}}(\hat{z})} \left[ \text{LCL}(\hat{z}) \right], \tag{4}$$

where $\gamma > 0$ is a hyperparameter to trade-off the balance between the two terms. In LCL, we use the latent variable $\hat{z}$, which is sampled from a distribution which we call *latent reference distribution*, and denote as $p_{\text{ref}}$. The latent reference distribution is the source of the data augmentation in latent space, where augmented latent variables are denoted with the $\widehat{\text{hat}}$ notation. We optimize the LCL of the $\hat{z}$ sampled from $p_{\text{ref}}$, which provides us versatility to shape the latent space at the region of interest and improve the sample-efficiency of the model. We discuss strategies to select $p_{\text{ref}}$ and their impact on latent space in §3.2.1.

The implementation of the LCA-VAE is simple; we can optimize (the parameters of) encoder $f_\phi^{\text{enc}}$ and decoder $f_\theta^{\text{dec}}$ of the VAE using an empirical version of the objective function in (4), where an extra term

$$\frac{1}{N^*} \sum_{i=1}^{N^*} ||\hat{z}_i - \hat{z}_i^1||^2 \simeq \mathbb{E}_{\hat{z} \sim p_{\text{ref}}(\hat{z})} \left[ \text{LCL}(\hat{z}) \right], \tag{5}$$

is subtracted (with multiplication with hyperparameter $\gamma$) from the empirical objective function of the standard VAE in (1). Here, $\hat{z}_i, i \in [N^*]$, is a latent variable sampled from latent reference distribution $p_{\text{ref}}$, and $N^*$ is the number the sampled latent variables.

### 3.2.1 Data Augmentation in Latent Space

In the LCA-VAE, where to augment new latent variables are determined by setting the parameters of the reference distribution, $p_{\text{ref}}$, and it can be customized based on the target task. In LSBO tasks, a common approach is to begin with a pretrained VAE and then refine it using newly acquired information through each step of BO. Consequently, we consider two types of reference distributions: one for the initial pretrained phase and another for the subsequent BO phase.

In the pretraining phase, our goal is to promote latent consistency broadly throughout the latent space. Therefore, we consider the reference distribution $p_{\text{ref}}(z)$ such that it spans the entire support of the VAE's latent variables' prior distribution. Following this strategy, we adopt a spherical multivariate normal distribution for $p_{\text{ref}}$, expressed as $N(\boldsymbol{\mu}_{\text{ref}}, \sigma_{\text{ref}}^2 I)$, where $\boldsymbol{\mu}_{\text{ref}} \in \mathcal{Z}$ is the mean vector, and $\sigma_{\text{ref}} > 0$ is the standard deviation. If the prior distribution of a VAE is set as the standard normal distribution $N(\mathbf{0}, I)$, we choose $\sigma_{\text{ref}}$ to be larger than 1 to explore a wider range through augmented latent variables from the sparse regions in the latent space[9]. Figures 2(C) and (D) demonstrate the latent space of LCA-VAE with this latent reference distribution. Unlike Figs. 2(A) and (B), we observe an improvement in latent consistency (Fig. 2(C)), with increased coverage of the latent space in the expanded region of latent consistency (Fig. 2(D)).

For the BO phase, the objective shifts towards reinforcing latent consistency in regions identified as promising regions by the LSBO's AFs. Hence, the reference distribution $p_{\text{ref}}(\boldsymbol{z})$ is dynamically adjusted based on the prevailing AF during each BO cycle. Specifically, we set the $\boldsymbol{\mu}_{\text{ref}} \leftarrow \arg\max \hat{f}^{\text{LCA-AF}}(\boldsymbol{z}^B)$. This indicates that we perform data augmentation in latent space by sampling latent variables around the consistent point with the highest AF value, aiming to bolster latent consistency in that region. The decision to use AF values as the primary driver for data augmentation, rather than other potential metrics, stems from the objective of targeting regions in the latent space that are most likely to yield the desired de-novo instances. Given that AFs are integral tools in the BO that are used to quantify the potential of regions within the search space, they present a logical and direct choice for guiding data augmentation in latent space. Thus, AF values become instrumental in updating the VAE. By iteratively expanding and refining the augmented search space in the latent space, we expect an elevated likelihood of unveiling desired de-novo instances.

**Preference of Data Augmentation in Latent Space over Input Space:** The choice between data augmentation in latent space and input space can be questioned since the latter is a well-studied methodology that has proven to be useful in many settings. In our context, data augmentation in input space refers to the method where new instances sampled from the latent space are directly appended to the original training instances. Therefore, there appears a pivotal question: Why prefer data augmentation in latent space over data augmentation in input space?

Augmentations in the input space, while intuitively appealing, introduce challenges during the VAE update process in LSBO. Specifically, these challenges relate to the representation and reconstruction of instances. When augmented instances are generated from sparsely populated regions of the latent space, the resultant outputs often diverge from the original input instance distribution. This divergence is problematic because, when the VAE attempts to reconstruct these instances, it is often challenging to simultaneously train both the original input instances and highly different newly generated instances. This misalignment gives rise to exacerbated reconstruction errors. Such errors are further complicated by the expansive and high-dimensional nature of the input space, commonly referred to as the *curse of dimensionality*. This complexity inflates the scale of reconstruction errors, making the optimization landscape more challenging to navigate. These challenges pose the risk of encountering a phenomenon known as catastrophic forgetting, as described by McCloskey and Cohen

---

[9]In our numerical experiments in §4, we set $\sigma_{\text{ref}}$ to 2.

[24]. In such cases, the models inadvertently overwrite previously well-learned reconstructions, thereby undermining their overall performance and reliability. Finally, the data augmentation in input space does not inherently address the pervasive issue of latent inconsistencies, which causes limitations in the exploratory capabilities of LSBO algorithms.

Conversely, data augmentation in latent space offers significant advantages. The latent space enables highly targeted and accurate augmentations within the LSBO framework. Inherently, the latent space is computationally more tractable and adheres to a standard normal distribution. This feature effectively mitigates the problem of amplified errors, in this case the LCL values, thereby do not over-complicating the optimization process. Another benefit lies in its ability to address latent inconsistencies when integrated with the LCL within the LCA-VAE framework, fortifying latent consistency particularly within promising regions in the latent space. This aids LSBO tasks by efficiently leveraging label-free instances, accelerating the evolution of the VAE after each retraining step.

### 3.3 LCA-LSBO

Combining LCA-AF and LCA-VAE, we propose LCA-LSBO for improved LSBO performance. LCA-LSBO harnesses the consistent points in the latent space using LCA-AF and expands their presence by integrating data augmentation in latent space with LCL, providing targeted consistency in region of interest via LCA-VAE. Algorithm 3 describes LCA-LSBO. Starting from the pretrained LCA-VAE, we use the LCA-AF to pinpoint the highest value query point. Next, we set the latent reference distribution's center $\boldsymbol{\mu}_{\text{ref}}$ to that point, and set $\sigma_{\text{ref}} < 1$ to focus the data augmentation in latent space, $\hat{\boldsymbol{z}}$, in the close vicinity of the region where AF is maximized[10]. We draw $\{\hat{\boldsymbol{z}}_i\}_{i \in [N^*]}$ using such $p_{\text{ref}}$, and retrain the LCA-VAE model. Once we reach the retraining iteration limit, we obtain $\hat{\boldsymbol{x}}^* = f_\theta^{\text{dec}}(\boldsymbol{\mu}_{\text{ref}})$ and $y^* = f^{\text{BB}}(\hat{\boldsymbol{x}}^*)$, subsequently adding a new instance $(\boldsymbol{\mu}_{\text{ref}}, y^*)$ to the labeled training set for the GP surrogate model in the latent space.

## 4 Numerical Experiments

We conducted three distinct LSBO tasks. The first task involves generating novel fashion products using the Fashion MNIST dataset. The second task focuses on creating new digits from scratch, utilizing the MNIST dataset. We present the first two tasks in §4.1. Our final task focuses on de-novo chemical compound design, aiming to identify new compounds that achieve the optimal docking scores when interacting with a specific protein. We present the final task in §4.2.

### 4.1 De-Novo Image Generation

The literature on de-novo instance generation has primarily focused on and found its implementation in the domain of chemical design. While following this common phenomenon, we argue that de-novo chemical design in LSBO literature faces a fundamental challenge: the difficulty of interpreting the resulting chemical compounds beyond their evaluation through the BB function. Even though LSBO can discover novel instances with desired scores (such as desired docking scores), the resulting chemical compounds may often be far from stable and synthesizable. Hence, we strongly endorse the adoption of an LSBO experimental framework that prioritizes objectives with enhanced interpretability. In response to this requirement, we have developed an LSBO experimental framework that capitalizes on two widely recognized datasets in the machine learning literature: Fashion MNIST and MNIST datasets. In the following sections, we will elucidate the proposed LSBO experiment setup using these datasets and delve into the corresponding results.

---

[10]Specifically, we set $\sigma_{\text{ref}} = 0.3$ in the LCA-LSBO experiments in §4.

**Algorithm 3** Latent Consistency Aware-LSBO
**Require:** Unlabeled instances: $\mathcal{U} = \{x_i\}_{i \in [\mathcal{U}]}$, Labeled instances: $\mathcal{L} = \{(x_i, y_i)\}_{i \in [\mathcal{L}]}$, $f^{\text{AF}}(z)$, AF $f^{\text{AF}}$, BB Function $f^{\text{BB}}$, Experiment count $J$, Burn-in threshold $B$, Standard Deviation of $p_{\text{ref}}$: $\sigma_{\text{ref}}$, Sample size of $\hat{z}$: $N^*$
1: Train LCA-VAE with $p_{\text{ref}}$ $\mathcal{N}(\mathbf{0}, 2 * I)$
2: **for** $j = 1$ to $J$ **do**
3:     Fit GP model using $\{(f_\phi^{\text{enc}}(x_i), y_i)\}_{i \in \mathcal{L}}$
4:     Find $z \leftarrow \arg\max_{z \in \mathcal{Z}} \hat{f}^{\text{LCA-AF}}(\hat{z}_B)$ and set $\boldsymbol{\mu}_{\text{ref}} \leftarrow \hat{z}_B$
5:     Generate object: $\hat{x}^* = f_\theta^{\text{dec}}(z)$
6:     Evaluate label: $y^* = f^{\text{BB}}(\hat{x}^*)$
7:     Update: $\mathcal{L} \leftarrow \mathcal{L} \cup (\hat{x}^*, y^*)$
8:     Obtain samples $\{\hat{z}_i\}_{i \in [N^*]} \sim p_{\text{ref}}$
9:     ReTrain LCA-VAE using $\{\{x_i\}_{i \in [\mathcal{U}]} \cup \{x_i\}_{i \in [\mathcal{L}]}\}$ and $\{\hat{z}_i\}_{i \in [N^*]}$
10: **end for**

### 4.1.1 Experiment Setup

Generative models have been trained effectively on the Fashion MNIST and MNIST datasets, consistently achieving good performance. Nonetheless, by introducing a slight modification, the generative task's difficulty can be amplified.

*Can we generate "bag" using a VAE trained without "bag" instances?*

The Fashion MNIST dataset encompasses ten distinct fashion categories: t-shirt/top, trouser, pullover, dress, coat, sandal, shirt, sneaker, ankle boot, and *bag*[11]. Consider training a VAE model on nine of these categories, intentionally excluding instances labeled as *bag*, as demonstrated in Fig. 4(A). Given the absence of exposure to *bag*-labeled instances during its training, generating bag-like instances becomes a challenge for the VAE. This scenario falls under the LSBO framework, aiming to generate a de-novo instance, specifically a *bag*, which is a unique entity absent from the model's training instances.

---
[11]Example of Fashion MNIST instances are provided in Appendix.

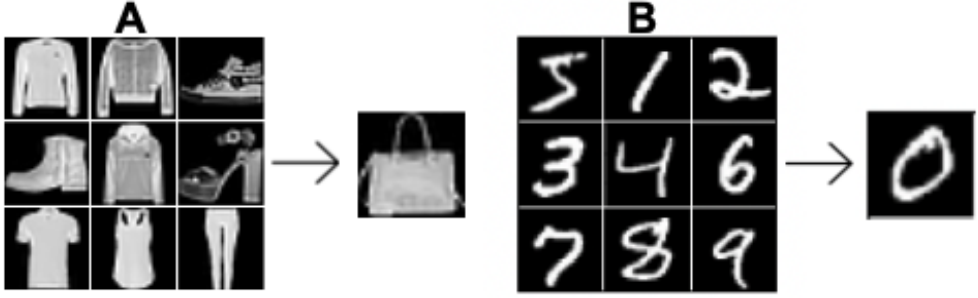

**Fig. 4** Demonstration of the input instances and target instances in Fashion MNIST and MNIST experiments in §4.1. **A** shows the input instances of Fashion MNIST VAE, which does not include bag-labeled instances, and the target *bag* instance to be generated from such VAE. **B** shows the input instances, which do not include the digit *0*, for MNIST VAE with the task of generating digit *0*.

### Can we generate "0" using a VAE trained without "0" instances?

A similar setting can be created for the MNIST dataset as well. Consider training a VAE with MNIST digits $D \in \{1, 2, 3, 4, 5, 6, 7, 8, 9\}$, intentionally excluding the digit 0, as demonstrated in Figure 4(B). Next, such a VAE is challenged to generate the excluded digit 0. As in the Fashion MNIST *bag* generation task, this is an LSBO scenario since the objective is to generate a de-novo instance, specifically digit 0, which is not included in the model's training instances.

Our study extended across all classes in Fashion MNIST and digits in MNIST. Ten categories per dataset led to ten LSBO tasks for each, where a VAE trained without the excluded class/digit and then challenged to generate the excluded class/digit. This approach rigorously tests the de-novo generation abilities of different LSBO methodologies through 20 distinctive tasks.

Addressing these problems within the LSBO framework necessitates a BB function for assessing instances produced during LSBO iterations. To mimic such a BB function, a classifier was employed to estimate the probability that a generated image represents the target image. Therefore, set of classifiers, serving as the BB function, was trained to discern the target instance from others. For the *bag* generation task explained above, a classifier was employed to estimate the probability that a generated image represents a *bag*. For the "0" generation task explained above, a classifier was employed to estimate the probability that a generated image represents "0" to mimic the BB function. These classifier's probability score for an image being the target instance defines the BB function's output[12]. The same setting is created for other de-novo image generation experiments as well. The primary goal is to produce an image that maximizes the output of the BB function, with sample-efficiency as a core principle. Through these LSBO tasks, we showcase the de-novo object generation performance of our approach and highlight the constraints of other existing methods.

---

[12]For each classifier, a neural network based model was trained by labeling target instances as 1 and the remaining instances in the corresponding dataset as 0.

### 4.1.2 Models

We examined three versions of the proposed method. They are called LCA-AF, LCA-AF(RT), and LCA-LSBO, where RT stands for "ReTraining". LCA-AF and LCA-AF(RT) refer to the LSBO method that uses the proposed LCA-AF as the AF in §3.1. In the former, we used a fixed VAE throughout the LSBO process, meaning that we did not retrain the VAE, In the latter, we retrained the VAE every time we obtained a new labeled instance. LCA-LSBO is a method that combines both the LCA-AF in §3.1 and the LCA-VAE in §3.2. The reason for considering not only the best-performing LCA-LSBO but also suboptimal LCA-AF and LCA-AF(RT) is to assess the impact of retraining the VAE and our two main proposals: focusing only on latent consistent points during BO and retraining with the LCA-VAE method. By isolating these components, we aim to gain a more detailed understanding of how each one contributes to the overall performance.

As baselines, we considered 9 methods: vanilla-VAE, vanilla-VAE(RT), pred-VAE, pred-VAE(RT), cond-VAE, cond-VAE(RT), weighted RT, DML RT, and LOL-BO. The first two baselines: vanilla-VAE and vanilla-VAE(RT) refer to the standard LSBO methods using a vanilla VAE [1] without or with retraining option, respectively. The next two baselines pred-VAE and pred-VAE(RT) refer to the LSBO methods with predictor [1] without or with retraining option, respectively. The next two baselines cond-VAE and cond-VAE(RT) refer to the LSBO methods in which conditional VAEs are used without or with the retraining option, respectively. The next baseline weighted RT [16] is an LSBO method in which the instances are weighted according to their label values when the VAE is retrained. Other baseline, DML RT [17], is an LSBO method in which the distance metric learned in the latent space is incorporated when the VAE is retrained. The last baseline, LOL-BO [3], is an LSBO method that uses local search by using the idea of Trust Region BO (TURBO) [25] in the LSBO framework.

In all versions of the proposed method and baseline methods except cond-VAE and cond-VAE(RT), the dimension of the latent spaces was set to 2. Since cond-VAE and cond-VAE(RT) require additional latent dimension, the dimension of their latent spaces was set to 3. For a detailed discussion of hyperparameter selection, see Appendix.

### 4.1.3 Results

We began by exploring baseline methods that do not incorporate retraining of the VAE with newly obtained instances during the BO process. For each method and task, we conducted BO for a total of 200 iterations in both Fashion MNIST and MNIST experiments. The outcomes of these experiments are shown in Table 1 and Table 2, respectively, where the best results after every $40^{th}$ iteration are provided for reference.

Interestingly, we observed that for each experiment, all the methods without retraining (including LCA-AF) completely failed to generate the unseen target bag and digit instances. This observation highlights the concept we initially proposed regarding the mismatch between the objectives of the VAE and the BO. Essentially, the VAE is designed to generate instances similar to its training data, limiting its ability to produce entirely new instances.

In subsequent Fashion MNIST and MNIST de-novo image generation experiments, we assessed the impact of retraining the VAE with new data from the LSBO process. In order to evaluate their sample-efficiency, we limited the number of LSBO steps to only 10 and retrained the model after each step. Table 3 and Table 4 show the generated samples with the highest probability of being a target instance for Fashion MNIST and MNIST experiments, respectively, where each image corresponds to the current best after every second experiment. The results demonstrate that performance improved with VAE retraining.

For Fashion MNIST experiments, starting from the *bag* generation task, we observed that competitor methods tended to generate off-target images. Among all competitor methods, the results demonstrate that pred-VAE(RT) seemed to provide images most resembling a bag, while it failed to add the holder of the *bag*. On the other hand, LCA-AF(RT) provided better performance compared to all the baselines, with generated instances bearing a resemblance to a bag. Furthermore, LCA-LSBO demonstrated the best performance among all, where the model could learn to generate bag-like instances. *Bag*-like images began to appear after $4^{th}$ retraining, which means only 4 queries of the BB function, and near-to-perfect results were achieved after $6^{th}$ retraining, showcasing the superior performance of LCA-LSBO. Furthermore, LCA-LSBO is also successful in all remaining tasks, it can generate (*T-shirts, Trousers, Pullovers, Dresses, Coats, Sandals, Shirts, Sneakers, Ankle Boots*) after a few retraining iterations, demonstrating the best performance among all methods. None of the competitor methods can achieve the similar performance. Among the competitor methods, LOL-BO is the only method that can generate *Ankle Boot*, while it fails to generate (*T-shirt, Dress, Trouser, Bag*). LCA-AF(RT) successfully generates (*Coat, Shirt, Sneaker*), showcasing better performance than vanilla-VAE(RT) in (*Bag, T-shirt, Coat, Sneaker*) tasks, demonstrating the advantage of focusing on latent consistent points during LSBO. Weighted RT and DML RT perform similarly, with a slight improvement in image quality in favor of DML RT. Finally, pred-VAE(RT) and cond-VAE(RT) performs similarly in generation of (*Bag, Pullover, Coat, Shirt*).

For MNIST experiments, we observed that vanilla-VAE(RT) is successful at two tasks (*1,6*), pred-VAE(RT) was successful at three tasks (*1,5,7*), and cond-VAE(RT) was successful at three tasks (*0,6,9*). Weighted RT improved upon vanilla-VAE(RT), while DML RT could identify the same digits as Weighted RT and provided better performance for digit *2*. LOL-BO method managed to be successful in the generation of digits (*1,5,6*), while it provided approximate representations of the digits (*0,9*). On the other hand, we observed that LCA-AF(RT) provided better performance compared to all the baselines: it could successfully generate the target digit in five out of ten tasks (*0,1,2,3,6,7*), while approximating the digit *8*. This suggests that focusing on latent consistent points via LCA-AF can significantly enhance LSBO performance. Finally, LCA-LSBO exhibited superior performance compared to all the other methods: it successfully accomplished all ten tasks, demonstrating that combining LCA-AF and LCA-VAE provides improvement.

The successful results in Fashion MNIST and MNIST experiments demonstrate that by using data augmentation in latent space and their proposed combination with LCL, LCA-LSBO can better populate and explore the latent space, update it in a sample-efficient way and identify the object of interest. This highlights the potential of our proposed method in effectively generating target objects, thanks to its improved exploration capabilities in the latent space.

**Table 1** Performance of baseline methods that do not use retraining.

| Class | vanilla-VAE | pred-VAE | cond-VAE | LCA-AF |
|---|---|---|---|---|
| Bag | | | | |
| T-shirt | | | | |
| Trouser | | | | |
| Pullover | | | | |
| Dress | | | | |
| Coat | | | | |
| Sandal | | | | |
| Shirt | | | | |
| Sneaker | | | | |
| Ankle Boot | | | | |

**Table 2** Performance of baseline methods that do not use retraining.

| Digit | vanilla-VAE | pred-VAE | cond-VAE | LCA-AF |
|---|---|---|---|---|
| 0 | | | | |
| 1 | | | | |
| 2 | | | | |
| 3 | | | | |
| 4 | | | | |
| 5 | | | | |
| 6 | | | | |
| 7 | | | | |
| 8 | | | | |
| 9 | | | | |

**Table 3** Each image represents the image with highest probability of being the target class after every second retraining.

| Class | vanilla-VAE(RT) | pred-VAE(RT) | cond-VAE(RT) | LOL-BO |
|---|---|---|---|---|
| Bag | | | | |
| T-shirt | | | | |
| Trouser | | | | |
| Pullover | | | | |
| Dress | | | | |
| Coat | | | | |
| Sandal | | | | |
| Shirt | | | | |
| Sneaker | | | | |
| Ankle Boot | | | | |
| **Class** | **weighted RT** | **DML RT** | **LCA-AF(RT)** | **LCA-LSBO** |
| Bag | | | | |
| T-shirt | | | | |
| Trouser | | | | |
| Pullover | | | | |
| Dress | | | | |
| Coat | | | | |
| Sandal | | | | |
| Shirt | | | | |
| Sneaker | | | | |
| Ankle Boot | | | | |

## 4.2 De-Novo Chemical Design

In this experiment, we utilized the ZINC250K dataset [1] to train our VAE. Each chemical compound in this dataset is represented using the SELFIES format [26], as it provides superior validity properties over the more commonly used SMILES representation [27]. Using SELFIES representation allowed us to train our models without introducing any grammatical rules to address validity concerns for generated chemical compounds ([28, 29]). We implement the Transformer LCA-VAE model with Transformer Encoder and Transformer Decoder, both of which use 8 heads, 6 layers, and a 32-dimensional latent space.

Our objective in this experiment is to optimize the docking scores of molecules, aiming for the lowest possible score when docking to site18 of the KAT1 protein.

**Table 4** Each image represents the image with highest probability of being the target digit after every second retraining.

| Digit | vanilla-VAE(RT) | pred-VAE(RT) | cond-VAE(RT) | LOL-BO |
|---|---|---|---|---|
| 0 | | | | |
| 1 | | | | |
| 2 | | | | |
| 3 | | | | |
| 4 | | | | |
| 5 | | | | |
| 6 | | | | |
| 7 | | | | |
| 8 | | | | |
| 9 | | | | |
| **Digit** | **weighted RT** | **DML RT** | **LCA-AF(RT)** | **LCA-LSBO** |
| 0 | | | | |
| 1 | | | | |
| 2 | | | | |
| 3 | | | | |
| 4 | | | | |
| 5 | | | | |
| 6 | | | | |
| 7 | | | | |
| 8 | | | | |
| 9 | | | | |

Docking scores are computed using Schrödinger's simulator [30]. To facilitate this, we first determine the 3D representation of the generated chemical compound utilizing Schrödinger's LigPrep software. Subsequently, the Glide software, also by Schrödinger, is used to calculate docking scores.

Given the potential time consumption of docking score computation—ranging from a few minutes to half an hour depending on the complexity of the chemical compound—as well as the possibility of failure in creating the 3D structure of the molecule or docking the protein, we put significant emphasis on sample efficiency in our experiment. To establish a benchmark for our methodology, we randomly select 50, 000 chemical compounds from the ZINC database [31] and evaluate their docking scores, comparing these against the scores generated by our method over 500 iterations of the BO algorithm using Algorithm 3. We further impose a constraint that both generated chemical compounds and 50, 000 chemical compounds from the ZINC database have molecular weights lower than 450 Da (Dalton), calculated by summing the weights of each atom in the molecule, respecting the *Lipinski's Rule of Five* [32], which states that molecules with a molecular weight over 500 Da becomes harder to synthesize.

### 4.2.1 Results

We employ Algorithm 3 in conjunction with the Transformer LCA-VAE and execute 500 iterations of BO. The results of this experiment are depicted in Fig. 5. Figure 5(A) demonstrates the lowest docking score in each iteration of BO and the lowest docking score[13] from 50, 000 instances from ZINC database, and Fig. 5(B) demonstrates the distribution of the docking scores of the molecules.

As shown in Fig. 5(A), our method succeeds in generating a de-novo molecule with a docking score lower than 50, 000 instances from the ZINC database, and it achieves this in fewer than 500 iterations. The molecule with the lowest docking score among the 50, 000 selected randomly from the ZINC database exhibits a docking score of $-8.8$. Our approach identifies molecules with docking scores as low as $-9.58$ within just 500 iterations. Some of the molecules that achieved docking scores lower than $-8$ are showcased in Figure 6. Besides, Table 3 shows the comparison of the Top 3 lowest docking scores obtained by 50, 000 molecules from the ZINC database and our method, where our method provides better scores in each level. It shows that our method can explore many de-novo chemical compounds of interest within a limited number of BO iterations. Besides, Fig. 5(B) shows that docking score distribution of LCA-LSBO instances is shifted towards a lower value compared with those from ZINC database.

Given the significant time and financial investment required to conduct docking simulations using highly accurate simulators, it becomes imperative to have a model that exhibits high sample-efficiency and good exploration capabilities. As our experiments with the Fashion MNIST and MNIST datasets have also demonstrated, our proposed method exhibits promising sample-efficiency characteristics. Results underscore the importance of focusing on consistent points through LCA-AF, and employing LCA-VAE with LCA-AF in LSBO tasks.

**Table 5** Comparison of TOP 3 molecules with lowest docking scores.

| Method | Sample Size | $1^{st}$ | $2^{nd}$ | $3^{rd}$ |
|---|---|---|---|---|
| ZINC Database | 50,000 | -8.80 | -8.66 | -8.51 |
| **Ours** | **500** | **-9.58** | **-9.54** | **-9.44** |

## 5 Conclusion

In conclusion, our research has focused on developing a novel approach for efficient LSBO in complex scenarios. Our experiments have demonstrated the promising characteristics of our method, showcasing its high sample-efficiency and effective exploration capabilities. The results obtained from our experiments with the Fashion MNIST and MNIST datasets and docking scores have validated the effectiveness of our approach. The emphasis on consistent points through LCA-AF, coupled with the use of data augmentation in latent space within the LCA-VAE framework for LSBO tasks, has been instrumental in attaining optimal solutions. These findings highlight the potential of our LCA-LSBO approach for real-world applications, where the efficient use of resources is paramount.

---

[13]The lower the docking score, the better.

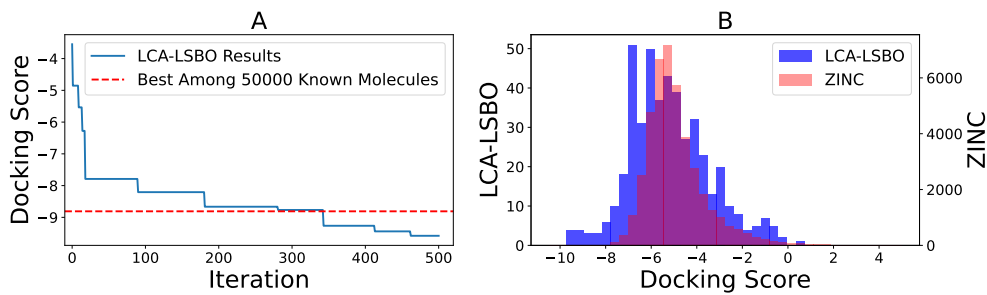

**Fig. 5** Comparison of LCA-LSBO and randomly selected 50,000 samples from ZINC database. **A** shows BO results of Docking Score experiments and the lowest docking score obtained by 50,000 samples. The lower, the better. LCA LSBO generates molecules with lower docking scores than 50,000 samples within 500 iterations. **B** shows the distribution of docking scores. This plot has two y-axes with different scales. The left y-axis of **B** demonstrates frequencies of de-novo instances found by LCA-LSBO and the right y-axis of **B** demonstrates frequencies of 50,000 ZINC samples. The distribution of instances found by LCA-LSBO is shifted towards lower values.

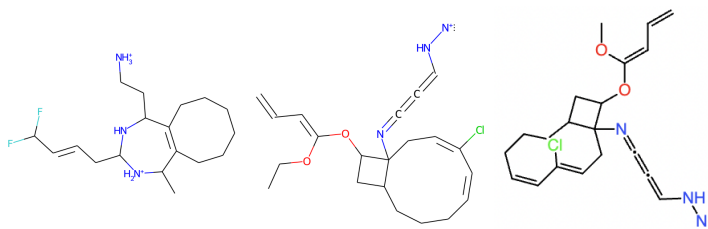

**Fig. 6** Example de-novo molecules found by our proposed method with docking scores values lower than $-8.8$.

## Acknowledgments

This work was partially supported by MEXT KAKENHI (20H00601), JST CREST (JPMJCR21D3, JPMJCR21D3), JST Moonshot R&D (JPMJMS2033-05), JST AIP Acceleration Research (JPMJCR21U2), NEDO (JPNP18002, JPNP20006) and RIKEN Center for Advanced Intelligence Project.

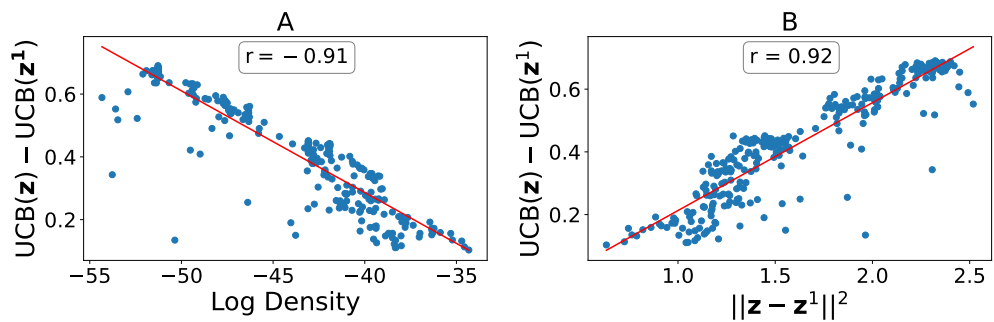

**Fig. 7** AF values, and squared differences of $z$ and $z^1$ using 32-dimensional latent space of VAE with Transformer Encoder and Transformer Decoder [33] trained on a database of chemical compounds, with UCB as AF. **A** shows the differences in the AF values against the density of $z$, **B** shows the differences in the AF values against squared distances of $z$ and $z^1$. Results indicates high UCB($z$) − UCB($z^1$) in low-density regions, and high correlation between UCB($z$) − UCB($z^1$) and $||z − z^1||^2$.

# Appendix

### List of Abbreviations

| Term | Abbreviation |
| :---: | :---: |
| Variational Autoencoder | VAE |
| Bayesian Optimization | BO |
| Retraining | RT |
| Acquisition Function | AF |
| Latent Space Bayesian Optimization | LSBO |
| Latent Consistent Aware | LCA |
| Latent Consistent Aware Latent Space Bayesian Optimization | LCA-LSBO |
| Latent Consistent Aware Acquisition Function | LCA-AF |
| Latent Consistent Aware Variational Autoencoder | LCA-VAE |
| Latent Consistency Loss | LCL |
| Gaussian Process | GP |
| Black-Box | BB |
| Kullback Leibler | KL |

**Table 6** List of abbreviations.

### Impact of Latent Incosistencies to AF Values

Figure 7(A) shows the relationship between the LCL and AF values. Figure 7(B) illustrates the changes in AF values with the degree of latent inconsistency $||z − z^1||^2$ when using the well-known AF called Upper Confidence Bound (UCB) in LSBO. It is evident that significant differences exist in the AF values within the regions of latent inconsistency. Consequently, even when the BO algorithm queries points with high AF values, it may end up querying points with low AF values.

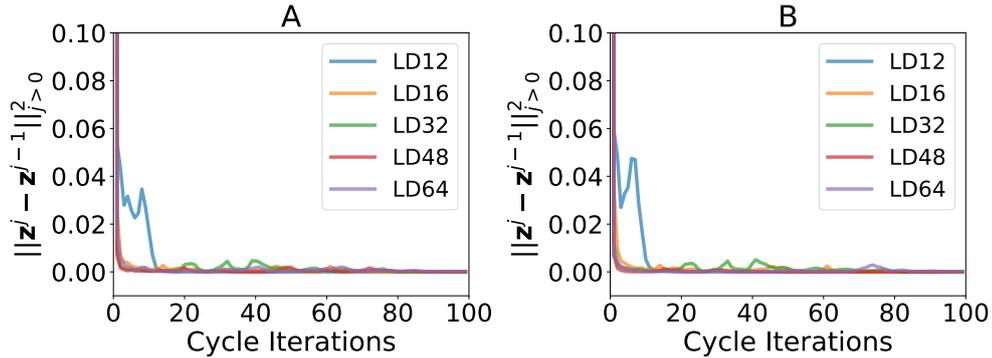

**Fig. 8** Illustration of cycle iterations and the variations between successive cycles for latent variables, sampled from 3 (**A**) and 4 (**B**) standard deviations away from the prior distribution of the VAEs with different latent dimensions. Although higher latent dimensions necessitate an increased number of iterations until convergence to a consistent point, we observe that convergence is consistently achieved.

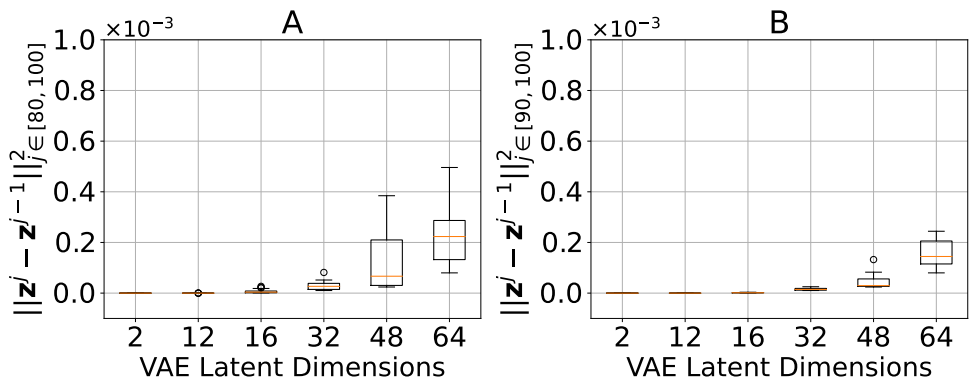

**Fig. 9** Distribution of error between successive cycles using the same initial $z$. **A** shows the distribution of the error values using $[z^{80}, z^{100}]$, and **B** shows the distribution of the error values using $[z^{90}, z^{100}]$. Note that the error values in y-axes are scaled by $10^{-3}$ in both figures. We observe that the magnitude of the error values is influenced by the latent dimension, and it decreases as the number of cycle iterations increases, taking negligible values.

## Consistency in High-Dimensional Latent Spaces

In §2.3.1, we explored how the sampling region of latent variables influences the number of iterations needed for convergence. Additionally, we found that the dimensionality of the latent space also significantly affects the convergence process. Specifically, as the dimensionality of the latent space increases, the number of iterations required to reach convergence increases, and the variation between successive iterations becomes more pronounced, especially when compared to lower-dimensional latent spaces.

To demonstrate these effects, we conducted experiments using VAEs with latent space dimensions set to 12, 16, 32, 48, and 64, each trained on the MNIST dataset. Our investigation focused on the number of iterations required to reach convergence for these VAEs, with latent variables sampled at distances of 3 and 4 standard deviations from the origin, in line with the VAEs' prior distribution. The outcomes, depicted in Figure 8, reveal a clear trend: the number of iterations needed for convergence increases alongside the dimensionality of the latent space. However, convergence remains achievable within a feasible number of iterations. Notably, the VAE with a 12-dimensional latent space reached convergence more swiftly than its counterparts, whereas the VAE with a 64-dimensional latent space required more iterations to achieve convergence.

On the other hand, the variation between consecutive cycle iterations tends to be more pronounced in higher-dimensional latent spaces than in those with fewer dimensions. To highlight the differences in error magnitudes, Figure 9 presents the error distributions for successive cycles in VAEs with latent dimensions of 2, 12, 16, 32, 48, and 64. Specifically, Figure 9(A) depicts the error distribution for cycled latent variables between $[z^{80}, z^{100}]$, while Figure 9(B) details the errors for cycles between $[z^{90}, z^{100}]$, where $z$ denotes a is a latent variable that is 3 standard deviations away from the origin. A clear pattern is observed in both figures: the error tends to increase with the dimensionality of the VAE's latent space. Notably, in both Figure 9(A) and (B), the VAE with a 2-dimensional latent space shows error values equal to zero. The VAEs with 12 and 16-dimensional latent spaces reach zero error values at later cycle iterations, as shown in Figure 9(B). A reduction in error values as we focus on the errors for cycles between $[z^{90}, z^{100}]$ is also noted in the remaining VAE configurations. The most significant errors are found in the 64-dimensional latent space VAE in both figures. While we observe error values below 0.0005 in Fig. 9(A), we observe a further decrease in error values in Fig. 9(B), with the peak error falling below 0.00035, which has a negligible impact on AF calculations.

Our analysis on the relationship between VAE's latent space, cycle iterations, and the magnitude of the error between successive cycles reveals that once the convergence point B is reached, the error between consecutive iterations beyond this point is consistently minimal and negligible. These findings enable us to simplify the LCA-AF calculation for VAEs with high-dimensional latent spaces as well. Therefore, we can set $\hat{f}^{\text{LCA-AF}}(z^j)_{j \in [B,M]} = f^{\text{AF}}(z^B)$.

## Optimization Details for LCA-AF

Let us denote the function that we cycle the latent variables as $T_B(z)$. This function encapsulates the composite effects of B decoding and encoding operations until convergence to the consistent point, resulting in $z^B = T_B(z)$. The AF is evaluated on these transformed variables, but the optimization process is aimed at finding the optimal $z$ that maximizes the AF evaluated on its transformed versions. With the simplification of the LCA-AF calculation, $\hat{f}^{\text{LCA-AF}}(z^j)_{j \in [B,M]} = f^{\text{AF}}(T_B(z))$, the optimization problem is formulated as:

$$z^* = \arg\max_{z} f^{\text{AF}}(T_B(z)). \qquad (6)$$

where $f^{\text{AF}}(T_B(z))$ represents the AF evaluated on the transformed latent variable $z^B$. The latent variable $z^*$ that maximizes this function indicates the most promising point for exploration in subsequent iterations, aligning with our objective to concentrate on consistent points within the latent space.

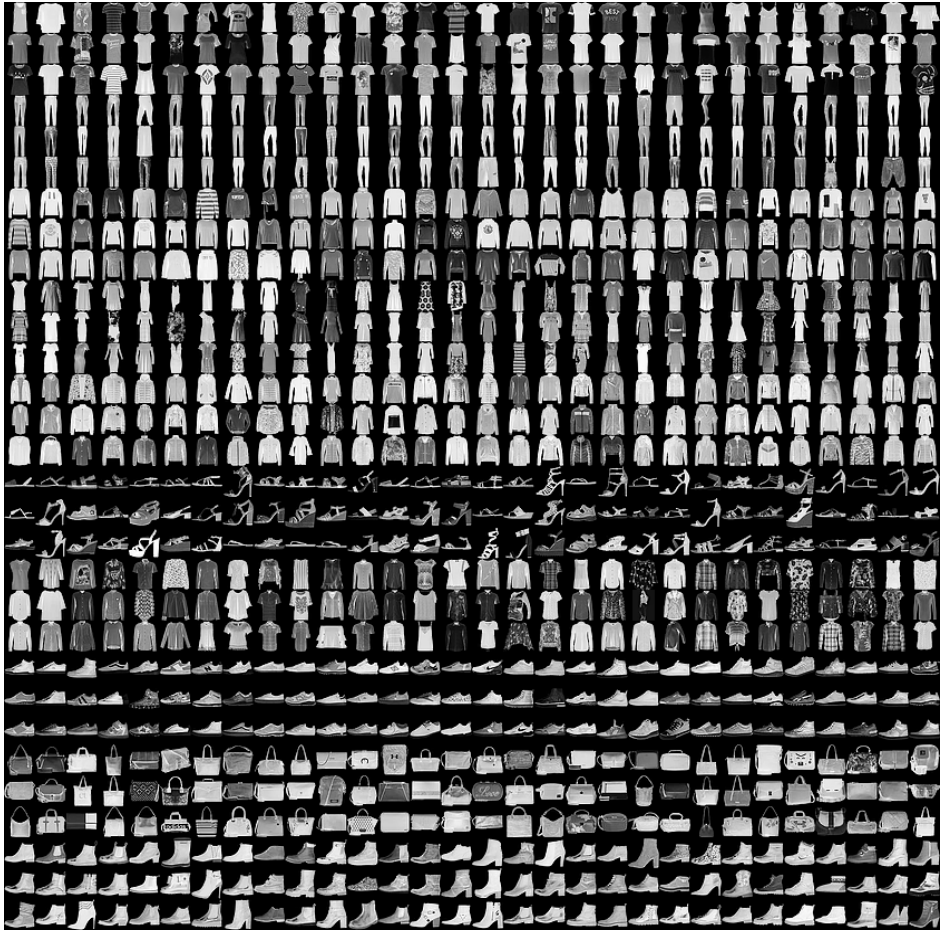

**Fig. 10** Types of images from 10 different instances. Each class has 3 different lines of images. The order of the classes from top to bottom are: T-shirt/top, trouser, pullover, dress, coat, sandal, shirt, sneaker, bag, ankle boot.

### Fashion MNIST Dataset Instances

In the Fashion MNIST dataset, there are different types of images within each class. Full set of example images are provided in the https://github.com/zalandoresearch/fashion-mnist, and we provide the same plot in Figure 10.

### Hyperparameter Optimization

In the proposed LCA-VAE method, we introduced the hyperparameter $\gamma$ to balance the weight of the LCL, and the $N^*$, the size of the data augmentation in latent space.

In order to optimize the $\gamma$ parameter in both the Fashion MNIST and MNIST experiments, we first trained vanilla-VAE models, and LCA-VAE models with $\gamma \in$ $[0.01, 0.1, 1, 10]$, where $N^*$ in each batch is set equal to the batch size of the training instances. For simplicity, LCA-VAEs are trained with all instances in their respective training datasets for both experiments, where the weight of the KL Divergence term, $\beta$, was set to 1 in each model. Upon training the models, we employed a method involving the generation of instances by sampling latent variables from a Uniform distribution within the range of [-6, 6]. These latent variables were then input into the decoder of models. In order to conduct hyperparameter optimization, we began by creating 100 sub-samples, each comprising 1000 instances randomly selected from the input dataset. In addition to this, we generated 100 sets of random instances using the models that are trained with respected $\gamma$ values, with each set containing 1000 samples, and these were generated using different random seeds. Once these datasets were prepared, we proceeded to employ them for the computation of Frechet Inception Distance (FID) [34] scores. For each pair of input instances and generated instances, we calculated the corresponding FID scores, and reported the mean and standard deviation values of the resulting FID scores. The primary objective was to assess the generative capabilities of the model across a wide span of the latent space.

Similarly, in the context of de-novo chemical design experiments, we conducted a comparative analysis between vanilla-VAE and LCA-VAE models using the same strategy. We trained a vanilla-VAE model and LCA-VAEs with $\gamma$ values $\gamma \in [0.01, 0.1, 1, 10]$, with $N^*$ in each batch equal to the batch size of the training instances. In this case, upon training the models, we assessed the diversity of the 100 sets of generated chemical compounds when new instances were generated by decoding from a Uniform[-6,6] distribution with a sample size of 1000, using 100 different seeds in order to report the mean and standard deviation of the percentage of the diversity values. Here, the diversity metric is calculated by dividing the number of unique chemical compounds by the total number of generations. Additionally, the $\beta$ value in VAE and LCA-VAE models in de-novo chemical design experiments was set to $\beta = 0.01$. This choice was influenced by a comprehensive analysis presented in [35], which empirically established that setting $\beta < 1$ leads to enhanced generative performance in chemical compound generation. Importantly, our own experiments further validated the efficacy of this choice.

The selection of the Uniform[-6,6] distribution for each experiment was motivated by the fact that BO does not adhere to a normal distribution during querying. Furthermore, the choice of [-6, 6] aligns with the search space typically associated with BO in LSBO literature [16, 17].

In Table 7, the results of the Fashion MNIST, MNIST, and chemical compound generation experiments are shown. Note that the row with $\gamma = 0$ represents the results from vanilla-VAE experiments. Our findings indicate that setting $\gamma = 0.01$ yielded the most favorable FID scores. Consequently, this value was chosen as the optimal $\gamma$ parameter for the LSBO experiments in both Fashion MNIST and MNIST experiments. For chemical compound experiments, results indicate that setting $\gamma = 0.1$ yielded the most promising outcomes. Notably, setting $\gamma \leq 1$ consistently yielded improved results across different settings, while the setting $\gamma > 1$ led to diminished generative performance. Decreased generative performance when $\gamma = 10$ is due to high penalization of inconsistencies which comes at the expense of the generative performance of the model.

Table 7 FID scores for two image generation tasks and diversity scores of chemical compound generations.

| $\gamma$ | FID for MNIST | FID for Fashion MNIST | Diversity of Chemical Compounds |
|---|---|---|---|
| 0 | $80.7 \pm 1.63$ | $108.5 \pm 1.98$ | $92.6\% \pm 1.2\%$ |
| 0.01 | $\mathbf{75.7} \pm 1.4$ | $\mathbf{99.8} \pm 1.73$ | $93.8\% \pm 0.7\%$ |
| 0.1 | $82.1 \pm 1.5$ | $103 \pm 1.61$ | $\mathbf{96.2}\% \pm 0.4\%$ |
| 1 | $83.3 \pm 1.5$ | $113.25 \pm 2.04$ | $94.5\% \pm 0.7\%$ |
| 10 | $87.4 \pm 1.4$ | $125.6 \pm 2.61$ | $91.8\% \pm 0.7\%$ |